%% file: paper.tex
\documentclass[10pt,twocolumn,letterpaper]{article}

\usepackage{cvpr}
\usepackage{times}
\usepackage{epsfig}
\usepackage{graphicx}
\usepackage{amsmath}
\usepackage{amssymb}

\usepackage[pagebackref=true,breaklinks=true,letterpaper=true,colorlinks,bookmarks=false]{hyperref}

\cvprfinalcopy % *** Uncomment this line for the final submission

\def\cvprPaperID{****} % *** Enter the CVPR Paper ID here

\input{tools}

\ifcvprfinal\fi
\begin{document}

%%%%%%%%% TITLE
\title{Wavelet Convolutional Neural Networks}

\author{Shin Fujieda\\
The University of Tokyo, Digital Frontier Inc.\\
{\tt\small sfujieda@graphics.ci.i.u-tokyo.ac.jp}
\and
Kohei Takayama\\
Digital Frontier Inc.\\
{\tt\small ktakayama@dfx.co.jp}
\and
Toshiya Hachisuka\\
The University of Tokyo\\
{\tt\small hachisuka@ci.i.u-tokyo.ac.jp}
}

\maketitle
%\thispagestyle{empty}

%%%%%%%%% ABSTRACT
\input{abstract}

%%%%%%%%% BODY TEXT
\input{intro}
\input{related}
\input{method}
\input{experiments}
\input{discussion}
\input{conclusion}

{\small
\bibliographystyle{ieee}
\bibliography{references}
}

\end{document}

%% file: tools.tex
\usepackage{microtype}
\usepackage{amsfonts}

\usepackage{suffix}
\usepackage{ifthen}
\usepackage{listings}
\usepackage{paralist}
\usepackage{wrapfig}

\usepackage{algorithm,algpseudocode}
\MakeRobust{\Call}

\usepackage{color}
\definecolor{blackgreen}{rgb}{0.0,0.4,0.0}
\definecolor{gray}{rgb}{0.5,0.5,0.5}

\usepackage{bbm}

\usepackage[draft]{pdfcomment}  % for work in progress

\defineavatar{TODO}{color=red,open=true,opacity=0.5}
\defineavatar{SF}{author=Shin Fujieda,color=magenta,open=false,opacity=0.5}
\defineavatar{TH}{author=Toshiya Hachisuka,color=cyan,open=false,opacity=0.5}

\newcommand{\IGNORE}[1]{}

\newcommand{\Chapter}[2][]
{
	\chapter{#2}
	\ifthenelse{\equal{#1}{}}{}{\label{#1}}
}

\newcommand{\Section}[2][]
{
	\section{#2}
	\ifthenelse{\equal{#1}{}}{}{\label{#1}}
}

\newcommand{\SubSection}[2][]
{
	\subsection{#2}
	\ifthenelse{\equal{#1}{}}{}{\label{#1}}
}

\newcommand{\SubSubSection}[2][]
{
	\subsubsection{#2}
	\ifthenelse{\equal{#1}{}}{}{\label{#1}}
}

\WithSuffix\newcommand\SubSubSection*[2][]
{
	\subsubsection*{#2}
}

\newcommand{\Paragraph}[2][]
{
	\paragraph{#2}
	\ifthenelse{\equal{#1}{}}{}{\label{paragraph:#1}}
}

\newcommand{\SubParagraph}[2][]
{
	\subparagraph{#2}
	\ifthenelse{\equal{#1}{}}{}{\label{paragraph:#1}}
}

\newcommand{\Figure}[4][]
{
	\begin{figure}[htb]
		\begin{center}
			\includegraphics[width = \linewidth]{images/#2}\vspace*{-2mm}
            \ifthenelse{\equal{#4}{}}{}{\caption{#4}}
			\ifthenelse{\equal{#3}{}}{}{\label{#3}}
		\end{center}
	\end{figure}
}

\newcommand{\FigureFull}[4][]
{
	\begin{figure*}[#3]
		\begin{center}
			\includegraphics[width = \linewidth]{images/#2}
            \ifthenelse{\equal{#4}{}}{}{\caption{#4}}
			\ifthenelse{\equal{#1}{}}{}{\label{#1}}
		\end{center}
	\end{figure*}
}

\newcommand\Math[2][XXX]{%
	\begin{equation}%
		\ifthenelse{\equal{#1}{XXX}}{\nonumber}{\label{#1}}
		\begin{split}%
			#2%
		\end{split}%
	\end{equation}%
}

\newcommand\MathAlign[2][XXX]{%
	\begin{align}%
		\ifthenelse{\equal{#1}{XXX}}{\nonumber}{\label{#1}}
        #2%
	\end{align}%
}

\algnewcommand\Not{\textnormal{\textbf{not}}\xspace}

\definecolor{CommentColor}{rgb}{0,0.5,0}

\DeclareFixedFont{\mono}{T1}{fvm}{m}{n}{9pt}

\definecolor{BackgroundColor}{rgb}{1,1,1}
\definecolor{CodeColor}{rgb}{0,0,0}
\definecolor{KeyWordColor}{rgb}{0,0,0}
\definecolor{EmphasizeColor}{rgb}{0.15,0.15,0.35}

\newcommand{\SetDefaultListingParams}
{
	\lstset
	{
		captionpos=b,
		frame=none,
		language=C++,
		tabsize=2,
		lineskip=0pt,
		xleftmargin=0pt,
		xrightmargin=0pt,
        framexleftmargin=0pt,
		framexrightmargin=0pt,
        framextopmargin=0pt,
        framexbottommargin=0pt,
        columns=fullflexible,
		backgroundcolor=\color{BackgroundColor},
		basicstyle=\fontsize{7.6}{4}\sffamily\color{CodeColor},
		commentstyle=\color{CommentColor},
		keywordstyle=\bfseries\color{KeyWordColor},
		 emph={Pow, InitFrame,GetPaths,TraceRay,ProjectToEye,Render,RenderProgressive,TraceLightPaths,BuildRangeSearchStruct,IsValid,SamplePixel,Accumulate,Connect,Merge,SamplePointOnLight,SampleNextVertex,RangeSearch,TerminateRandomWalk,ContinueRandomWalk,IsEmissive,ComputeMeasurementContribution,ComputePDF,BalanceHeuristic},
        morekeywords={to,function,in},
		emphstyle={\color{EmphasizeColor}},
        mathescape=true
	}
}

\SetDefaultListingParams

\lstnewenvironment{Code}[3][]
{%
	\SetDefaultListingParams
	\ifthenelse
	{
		\equal{#1}{}
	}
	{
		\lstset
		{
			label=#2,
		}
	}
	{
		\lstset
		{
			label=#2,
			float=#1
		}
	}
	\lstset
	{
		caption={#3},
		nolol
	}
}{}

	\newenvironment{squishlist}
	{
		\begin{list}{$\bullet$}
		{
			\setlength{\itemsep}{0pt}
			\setlength{\parsep}{3pt}
			\setlength{\topsep}{3pt}
			\setlength{\partopsep}{0pt}
			\setlength{\leftmargin}{1.5em}
			\setlength{\labelwidth}{1em}
			\setlength{\labelsep}{0.5em}
		}
	}
	{
		\end{list}
	}

%% file: abstract.tex
%!TeX root=paper.tex

\begin{abstract}
Spatial and spectral approaches are two major approaches for image processing tasks such as image classification and object recognition.
Among many such algorithms, convolutional neural networks (CNNs) have recently achieved significant performance improvement in many challenging tasks.
Since CNNs process images directly in the spatial domain, they are essentially spatial approaches.
Given that spatial and spectral approaches are known to have different characteristics, it will be interesting to incorporate a spectral approach into CNNs.
We propose a novel CNN architecture, wavelet CNNs, which combines a multiresolution analysis and CNNs into one model.
Our insight is that a CNN can be viewed as a limited form of a multiresolution analysis.
Based on this insight, we supplement missing parts of the multiresolution analysis via wavelet transform and integrate them as additional components in the entire architecture.
Wavelet CNNs allow us to utilize spectral information which is mostly lost in conventional CNNs but useful in most image processing tasks.
We evaluate the practical performance of wavelet CNNs on texture classification and image annotation.
The experiments show that wavelet CNNs can achieve better accuracy in both tasks than existing models while having significantly fewer parameters than conventional CNNs.
\end{abstract}

%% file: intro.tex
%!TeX root=paper.tex

%%%%%%%%%%%%%%%%%%%%%%%%%%%%%%%%%%%%%%%%%%%%%%%%%%%%%%%%%%%%%%%
\section{Introduction}\label{Sec:Introduction}
%%%%%%%%%%%%%%%%%%%%%%%%%%%%%%%%%%%%%%%%%%%%%%%%%%%%%%%%%%%%%%%

\begin{figure*}[t!]
  \centering
	\includegraphics[width=\textwidth]{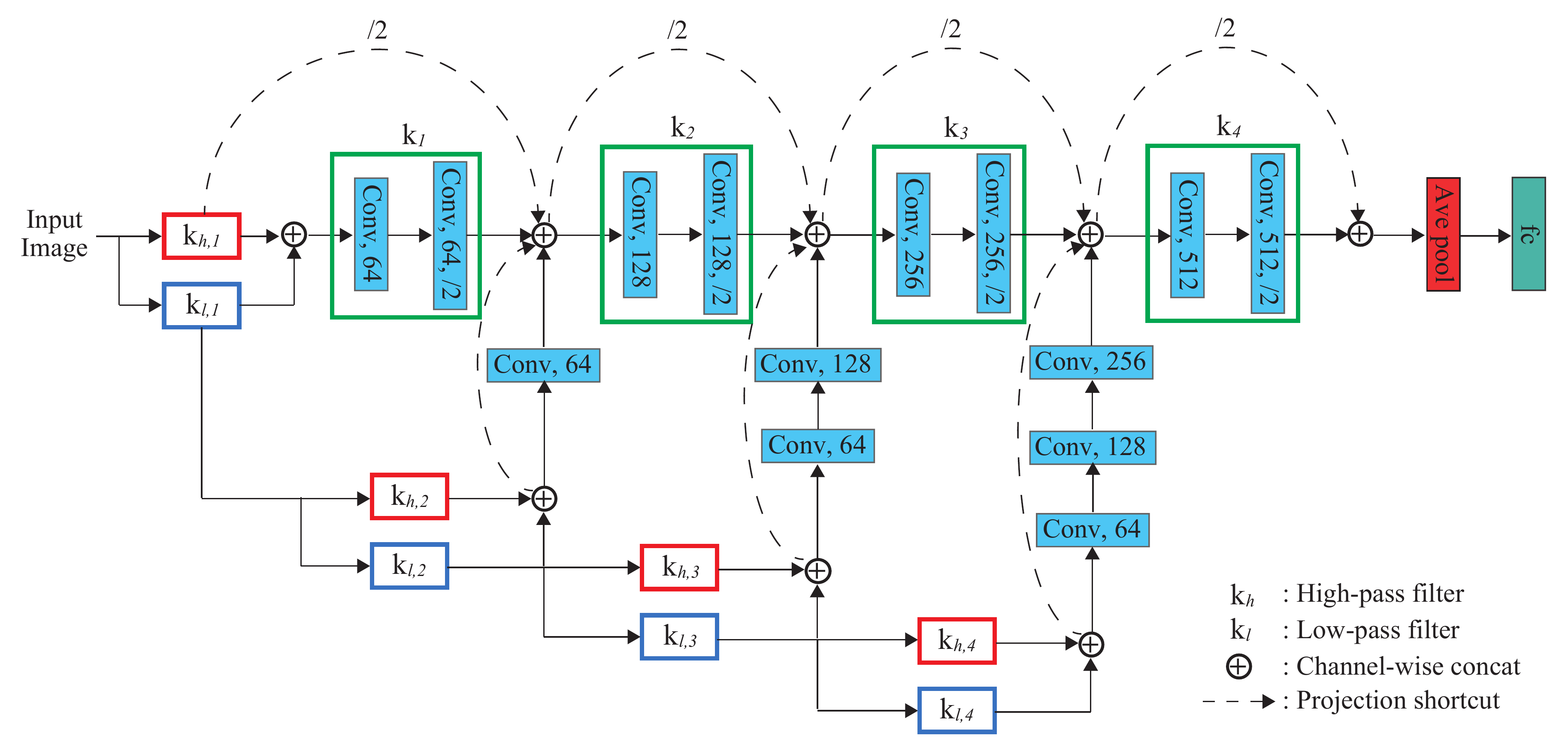}
  \vspace{-1em}
  \caption{Overview of wavelet CNN with 4-level decomposition of the input image. Wavelet CNN processes the input image through convolution layers with $3 \times 3$ kernels and $1 \times 1$ padding. The number after \emph{Conv} denotes the number of channels of the output. $3 \times 3$ convolutional kernels with the stride of $2$ and $1 \times 1$ padding are used to reduce the size of feature maps. Additionally, the input image is decomposed through multiresolution analysis and the decomposed images are concatenated channel-wise. The projection shortcuts are done by $1 \times 1$ convolutions. The output of convolution layers is vectorized by global average pooling followed by a fully connected layer (fc). The size of the output is equal to the number of classes included in the input dataset.}
  \vspace{-1em}
  \label{Fig:overview}
\end{figure*}

Convolutional neural networks (CNNs)~\cite{LeCun:1989,Krizhevsky:2012} are known to be good at capturing spatial features, while spectral analyses~\cite{Unser:1995,Lim:2010} are good at capturing scale-invariant features based on the spectral information.
It is thus preferable to consider both the spatial and spectral information within a single model, so that it captures both types of features simultaneously.
While the connection between CNNs and spectral approaches have been considered unclear so far, we found that a CNN can be seen as a limited form of a multiresolution analysis.
This observation points out that conventional CNNs are missing a large part of spectral information available via a multiresolution analysis.

We thus propose to supplement those missing parts of a multiresolution analysis as novel additional components in a CNN architecture.
%S
Figure~\ref{Fig:overview} shows the overview of our model; \emph{wavelet convolutional neural networks} (wavelet CNNs).
%S
Besides its theoretical formulation, we demonstrate the practical benefit of wavelet CNNs in two challenging tasks: texture classification and image annotation.
We demonstrate that wavelet CNNs achieve better or competitive accuracies with a significantly smaller number of trainable parameters than conventional CNNs.
Our model is thus easier to train, less prone to over-fitting, and consumes less memory than conventional CNNs.
To summarize, our contributions are:
\begin{squishlist}
	\item Combination of CNNs and a multiresolution analysis as one model.
	\item Reformulation of CNNs as a limited form of a multiresolution analysis.
	\item Accurate and efficient texture classification and image annotation using our model.
\end{squishlist}
%
%

%% file: related.tex
%!TeX root=paper.tex

%%%%%%%%%%%%%%%%%%%%%%%%%%%%%%%%%%%%%%%%%%%%%%%%%%%%%%%%%%%%%%%
\section{Related Work}\label{Sec:RelatedWork}
%%%%%%%%%%%%%%%%%%%%%%%%%%%%%%%%%%%%%%%%%%%%%%%%%%%%%%%%%%%%%%%
\paragraph{Convolutional Neural Networks:}
CNNs essentially replaced conventional hand-crafted descriptors such as the Bag of Visual Words (BoVW)~\cite{Csurka:2004} due to the superior performance in various tasks~\cite{Krizhevsky:2012}.
The original network architecture has been extended to deeper architectures since then.
One such architecture is Residual Networks (ResNets)~\cite{He:2016} which make deeper networks easier to train by introducing \emph{shortcut connections}; connections which skip a few layers and perform identity mappings.
Inspired by ResNets, various networks with shortcut connections have been proposed~\cite{Zagoruyko:2016,Zhang:2017,Huang:2016}.

Even with shortcut connections, however, deeper networks still have a problem that information about the input and gradient can quickly vanish as they propagate through the networks.
To address this problem, Dense Convolutional Network (DenseNet)~\cite{Gao:2017} further adds shortcut connections which connect each layer with all its previous layers.
While these networks have achieved impressive results on many computer vision tasks, they require significant computational resources because they have a large number of parameters.
We designed our network after DenseNet, but the combination with a multiresolution analysis allows us to significantly reduce the number of parameters compared to conventional networks.

Several recent works use CNNs as a feature extractor and a BoVW approach as pooling and encoding instead of the fully connected layers.
Cimpoi et al.~\cite{Cimpoi:2015} demonstrated that a CNN in combination with Fisher Vectors (FV-CNN) can achieve much better accuracy than using a CNN alone.
Their model uses a pre-trained CNN to extract image features and this CNN part is not trained with existing datasets.
Lin et al.~\cite{Lin:2017} achieved a remarkable improvement in fine-grained visual recognition by replacing the fully connected layers with bilinear pooling.
The dimension of the encoded features in the bilinear model is typically higher than 250,000 and it is difficult to train.
To address this difficulty, Gao et al.~\cite{Gao:2016} proposed compact bilinear pooling which reduces the number of parameters of bilinear pooling by 90\% while maintaining its performance.
Despite this significant reduction in the number of parameters, inherited from conventional CNNs, these models still have a large number of trainable parameters that makes them difficult to train in practice.
We show that our model achieves competitive results to compact bilinear pooling while further reducing the number of parameters in a texture classification task.

\paragraph{Spectral Approaches:}
Spectral approaches transform images into the frequency domain using a set of spatial filters.
The statistics of the spectral information at different scales and orientations define image features.
This approach has been well studied in image processing and achieved practical results~\cite{Unser:1995,Arivazhagan:2007,Kanchana:2013}.
Feature extraction in the frequency domain has an advantage.
A spatial filter can be easily made selective by enhancing certain frequencies while suppressing the others.
This explicit selection of certain frequencies is difficult to control in CNNs.
While CNNs are know to be universal approximators, in practice, it is unclear whether CNNs can learn to perform spectral analyses with available datasets.
Rather than relying CNNs to learn performing spectral analysis, we propose to directly integrate spectral approaches into CNNs, particularly based on a multiresolution analysis using wavelet transform~\cite{Mallat:1989}.
Our experiments show that a CNN with more parameters cannot be trained to become equivalent to our model with available datasets in practice.
%

%% file: method.tex
%!TeX root=paper.tex

%%%%%%%%%%%%%%%%%%%%%%%%%%%%%%%%%%%%%%%%%%%%%%%%%%%%%%%%%%%%%%%
\section{Wavelet Convolutional Neural Networks}\label{Sec:Method}
%%%%%%%%%%%%%%%%%%%%%%%%%%%%%%%%%%%%%%%%%%%%%%%%%%%%%%%%%%%%%%%

\paragraph{Overview:}
We propose to formulate convolution and pooling in CNNs as filtering and downsampling.
This formulation allows us to connect CNNs with a multiresolution analysis.
In the following explanations, we use a single-channel 1D data for the sake of brevity.
Applications to 2D images with multiple channels are trivially possible as was done by CNNs.

\subsection{Convolutional Neural Networks}
\begin{figure}[t]
  \centering
  \includegraphics[width=\columnwidth]{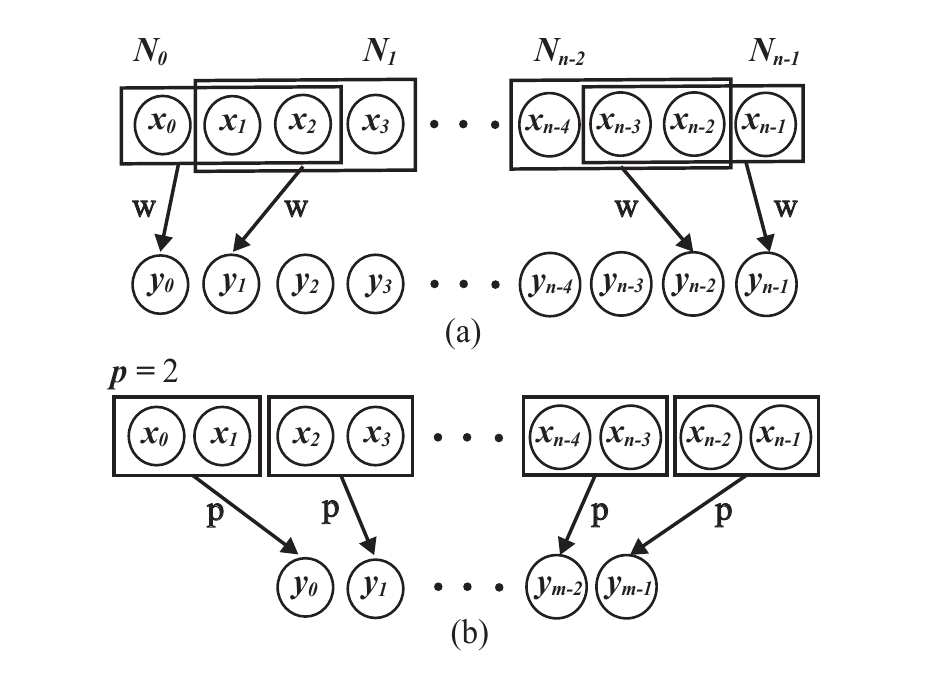}\vspace{-0.2cm}
	\vspace{-2mm}
	\caption{Concepts of convolution and average pooling layers. (a) Convolution layers compute a weighted sum of neighbor. (b) Pooling layers take an average and perform downsampling.}
	\vspace{-2mm}
	\label{Fig:GenPooling}
\end{figure}

In addition to the use of an activation function and a fully connected layer, CNNs introduce convolution/pooling layers.
Figure~\ref{Fig:GenPooling} illustrates the configuration we explain in the following.

\paragraph{Convolution Layers:}
Given an input vector with $n$ components $\mathbf{x} = (x_0, x_1, \ldots, x_{n-1}) \in \mathbb{R}^n$, a convolution layer outputs a vector of the same number of components $\mathbf{y} = (y_0, y_1, \ldots, y_{n-1}) \in \mathbb{R}^n$:
\Math[Eqn:Convolution]{
	y_{i} = \sum_{j \in N_i} w_{j} x_{j},
}
where $N_i$ is a set of indices of neighbors at $x_i$ and $w_{j}$ is a weight.
Following the notational convention in CNNs, we consider that $w_{j}$ includes the bias by having a constant input of $1$.
The equation thus says that each output $y_{i}$ is a weighted sum of neighbors $\sum_{j \in N_i} w_{j} x_{j}$ plus constant.

Each layer defines the weights $w_{j}$ as constants over $i$.
By sharing parameters, CNNs reduce the number of parameters and achieve translation invariance in the image space.
The definition of $y_{i}$ in Equation~\ref{Eqn:Convolution} is equivalent to convolution of $x_i$ via a filtering kernel $w_{j}$, thus this layer is called a convolution layer.
We can thus rewrite $y_{i}$ in Equation~\ref{Eqn:Convolution} using the convolution operator $\ast$ as
\Math[Eqn:Convolution2]{
	\mathbf{y} = \mathbf{x} \ast \mathbf{w},
}
where $\mathbf{w} = (w_0, w_1, \ldots, w_{o-1}) \in \mathbb{R}^o$.

\paragraph{Pooling Layers:}
Pooling layers are typically used immediately after convolution layers to simplify the information.
We focus on average pooling which allows us to see the connection with a multiresolution analysis.
Given an input $\mathbf{x} \in \mathbb{R}^n$, average pooling outputs a vector of a fewer components $\mathbf{y} \in \mathbb{R}^m$ as
\Math[Eqn:Pooling]{\vspace{-1EM}
	y_j =  \frac{1}{p} \sum_{k = 0}^{p-1} x_{p j + k},
}
where $p$ defines the support of pooling and $m = \frac{n}{p}$.
For example, $p = 2$ means that we reduce the number of outputs to a half of the inputs by taking pair-wise averages.
Using the standard downsampling operator $\downarrow$, we can rewrite Equation~\ref{Eqn:Pooling} as
\Math[Eqn:Pooling2]{
	\mathbf{y} = (\mathbf{x} \ast \mathbf{p} ) \downarrow p,
}
where $\mathbf{p} = (1/p, \ldots, 1/p) \in \mathbb{R}^p$ represents the averaging filter.
Average pooling mathematically involves convolution via $\mathbf{p}$ followed by downsampling with the stride of $p$.

\subsection{Generalized Convolution and Pooling}
Equation~\ref{Eqn:Convolution2} and Equation~\ref{Eqn:Pooling2} can be combined into a generalized form of convolution and downsampling as
\Math[Eqn:GenPooling]{
	\mathbf{y} = (\mathbf{x} \ast \mathbf{k} ) \downarrow p.
}
The generalized weight $\mathbf{k}$ is defined as
\begin{squishlist}
	\item $\mathbf{k} = \mathbf{w}$ with $p=1$ (convolution in Equation~\ref{Eqn:Convolution2})
	\item $\mathbf{k} = \mathbf{p}$ with $p>1$ (pooling in Equation~\ref{Eqn:Pooling2})
	\item $\mathbf{k} = \mathbf{w} \ast \mathbf{p}$ with $p>1$ (convolution followed by pooling).
\end{squishlist}
Our insight is that Equation~\ref{Eqn:GenPooling} is equivalent to a part of a multiresolution analysis.
To see this connection, let us consider convolution followed by pooling with $p=2$ and a pair of convolution kernels $\mathbf{k}_{l}$ and $\mathbf{k}_{h}$:
\Math[Eqn:Filtering]{
	&\mathbf{x}_l = (\mathbf{x} \ast \mathbf{k}_{l} ) \downarrow 2 \\
	&\mathbf{x}_h = (\mathbf{x} \ast \mathbf{k}_{h} ) \downarrow 2.
}
At this point, it is nothing but convolution and pooling with two different kernels $\mathbf{k}_{l}$ and $\mathbf{k}_{h}$.
The key idea is that a multiresolution analysis~\cite{crowley1981representation} decomposes $\mathbf{x}_l$ further as follows.

By defining $\mathbf{x}_{l, 0} = \mathbf{x}$, a multiresolution analysis performs a hierarchical decomposition of $\mathbf{x}_{l,t}$ into $\mathbf{x}_{l,t+1}$ and $\mathbf{x}_{h,t+1}$ by repeatedly applying Equation~\ref{Eqn:Filtering} with different $\mathbf{k}_{l,t}$ and $\mathbf{k}_{h,t}$ at each $t$:
\Math[Eqn:MRA]{
	&\mathbf{x}_{l,t+1} = (\mathbf{x}_{l,t} \ast \mathbf{k}_{l,t} ) \downarrow 2 \\
	&\mathbf{x}_{h,t+1} = (\mathbf{x}_{l,t} \ast \mathbf{k}_{h,t} ) \downarrow 2.
}
The number of applications $t$ is called a level in a multiresolution analysis.
Based on our reformulation, CNNs essentially discard $\mathbf{x}_{h,t}$ entirely and use only one set of kernels $\mathbf{k}_{t}$:
\Math[Eqn:MRA_CNN]{
	&\mathbf{x}_{l,t+1} = (\mathbf{x}_{l,t} \ast \mathbf{k}_{t} ) \downarrow 2.
}
Therefore, CNNs can seen as a limited form of a multiresolution analysis.

Figure~\ref{Fig:overview} illustrates how CNNs and our wavelet CNNs differ under this formulation.
We call $\mathbf{k}_{l,t}$ and $\mathbf{k}_{h,t}$ as low-pass and high-pass filter to follow the convention of multiresolution analyses.
Note, however, that they are not necessarily low-pass and high-pass filters in the spectral domain.
Conventional CNNs can be seen as a limited form of a multiresolution analysis that uses only $\mathbf{k}_{t}$ without a characteristic hierarchical decomposition (Equation~\ref{Eqn:MRA_CNN}).
Our model supplements the missing part due to $\mathbf{k}_{h,t}$ by introducing another set of $\mathbf{k}_{l,t}$ to form a multiresolution analysis via wavelet transform inside a neural network architecture.
While this idea might look simple after the fact, our model is powerful enough to outperform the existing more complex models as we will show in the results.

Note that we cannot use an arbitrary pair of filters ($\mathbf{k}_{l,t}$ and $\mathbf{k}_{h,t}$) to perform multiresolution analysis.
For wavelet transform, $\mathbf{k}_{h,t}$ is known as the wavelet function and $\mathbf{k}_{l,t}$ is known as the scaling function.
We used Haar wavelets~\cite{haar1910theorie} for our experiments, but our model is not restricted to Haar.
This constraint also suggests why it is difficult to train conventional CNNs to perform the same computation as wavelet CNNs do: weights $\mathbf{k}_{t}$ in CNNs are ignorant of this important constraint and just try to learn it from datasets.

Rippel et al.~\cite{Rippel:2015} proposed a related approach of replacing convolution and pooling by discrete Fourier transform and truncation of the coefficients.
This approach, called \emph{spectral pooling}, is equivalent to Equation~\ref{Eqn:MRA_CNN}, thus it is not essentially different from conventional CNNs.
Our model is also different from merely applying multiresolution analysis on input data and using CNNs afterward, since multiresolution analysis is built inside the network with skip connections.

\subsection{Implementation}
\paragraph{Network Structure:}
Figure~\ref{Fig:overview} illustrates our network structure.
We designed our main network structure after a VGG network~\cite{Simonyan:2014}.
We use $3 \times 3$ convolutional kernels exclusively and $1 \times 1$ padding to ensure the output is the same size as the input.

Instead of using the pooling layers to reduce the size of the feature maps, we exploit convolution layers with the increased stride.
If $1 \times 1$ padding is added to the layer with a stride of two, the output becomes half the size of the input layer.
This approach can be used to replace max pooling without loss in accuracy~\cite{Springenberg:2015}.
In addition, since both the VGG-like architecture and image decomposition in multiresolution analysis have the same characteristic that the size of images is reduced to a half successively, we combine each level of decomposed images with feature maps of the specific layer that are the same size as those images.

Furthermore, in order to use information of decomposed images more efficiently, we use \emph{dense connections}~\cite{Gao:2017} and \emph{projection shortcuts}~\cite{He:2016}.
Dense connections allow each level of decomposed images to be directly connected with all subsequent layers through channel-wise concatenation.
With this connectivity, our network can flow all the information effectively into the end of the network.
Projection shortcuts can be used to increase dimensions of the input with $1 \times 1$ convolutional kernels.
In our model, since dimensions of feature maps are different before and after the shortcut path, we use projection shortcuts in every shortcuts.
We also use \emph{global average pooling}~\cite{Lin:2014} instead of fully connected layers to prevent overfitting.
%
% Our model is implemented on Caffe~\cite{Jia:2014} and the codes for our model will be available on our website.

\paragraph{Learning:}
Wavelet CNNs exploit global average pooling with the same size as the input of the layer, so the size of input images is required to be the fixed size.
We thus train our proposed model exclusively with images of the size $224 \times 224$.
These images are achieved by first scaling the training images to $256 \times 256$ pixels and then conducting random crops to $224 \times 224$ pixels and flipping.
This random variation helps the model to prevent overfitting.
For further robustness, we use batch normalization~\cite{Ioffe:2015} throughout our network before activation layers during training.
For the optimizer, we exploit the Adam optimizer~\cite{Kingma:2014} instead of SGD.
We use the Rectified Linear Unit (ReLU)~\cite{Xavier:2011} as the activation function in all the experiments.
%

%% file: experiments.tex
%!TeX root=paper.tex
%%%%%%%%%%%%%%%%%%%%%%%%%%%%%%%%%%%%%%%%%%%%%%%%%%%%%%%%%%%%%%%
\section{Experiments}\label{Sec:Experiments}
%%%%%%%%%%%%%%%%%%%%%%%%%%%%%%%%%%%%%%%%%%%%%%%%%%%%%%%%%%%%%%%
We provide details of two applications of wavelet CNNs.
We applied wavelet CNNs to texture classification to confirm that wavelet CNNs can capture small features of images.
We also investigated how wavelet CNNs perform on natural images in an image annotation task.

\subsection{Texture Classification}
Texture classification is a challenging problem since textures often vary a lot within the same class, due to changes in viewpoints, scales, lighting configurations, etc.
In addition, textures usually do not contain enough information regarding the shape of objects which are informative to distinguish different objects in image classification tasks.
Due to such difficulties, even the latest approaches based on convolutional neural networks achieved a limited success, when compared to other tasks such as image classification~\cite{Hafemann:2014}.
Andrearczyk et al.~\cite{Andrearczyk:2016} proposed texture CNN (T-CNN) which is a CNN specialized for texture classification.
T-CNN uses a novel energy layer in which each feature map is simply pooled by calculating the average of its activated output.
This results in a single value for each feature map, similar to an energy response to a filter bank.
This approach does not improve classification accuracy, but its simple architecture reduces the number of parameters.

\paragraph{Datasets:}
For our experiments, we used two publicly available texture datasets: \emph{kth-tips2-b}~\cite{Hayman:2004} and \emph{DTD}~\cite{Cimpoi:2014}.
The \emph{kth-tips2-b} dataset contains 11 classes of 432 texture images.
Each class consists of four samples and each sample has 108 images.
Each sample is used for training once while the remaining three samples are used for testing.
The results for kth-tips2-b are shown as the mean and the standard deviation over the four splits.
The \emph{DTD} dataset contains 47 classes of 120 images "in the wild" which means that images are collected in uncontrolled conditions.
This dataset includes 10 available annotated splits with 40 training images, 40 validation images, and 40 testing images for each class.
The results for DTD are averaged over the 10 splits.
We processed the images in each dataset by global contrast normalization.
We calculated the accuracy as percentage of images that are correctly labeled which is a common metric in texture classification.

\begin{table}[t]
    \centering
    \scalebox{0.95}{
      \begin{tabular}{| l || l | l | l | l |} \hline
        & {2-level} & {3-level} & {4-level} & {5-level} \\ \hline
        kth-tips2-b & $59.1_{\pm 2.5}$ & $61.4_{\pm 1.7}$ & $63.5_{\pm 1.3}$ & ${\bf 63.7_{\pm 2.3}}$ \\ \hline
        DTD & $34.0_{\pm 1.6}$ & $35.2_{\pm 1.2}$ & ${\bf 35.6_{\pm 1.1}}$ & ${\bf 35.6_{\pm 0.7}}$ \\ \hline
      \end{tabular}
    }
    \caption{Classification results for different levels within wavelet CNNs trained from scratch indicated as accuracy (\%).}
    % \vspace{2mm}
    \label{Table:comparison_between_levels}
\end{table}

\begin{table}[t]
    \centering
    \begin{tabular}{| l || l | l || l |} \hline
      & \raisebox{0.1cm}{AlexNet} & \raisebox{0.1cm}{T-CNN} & \raisebox{-0.08cm}{\shortstack{Wavelet \\ CNN}} \rule[0mm]{0mm}{6mm}\\ \hline
      kth-tips2-b & $48.3_{\pm 1.4}$ & $49.6_{\pm 0.6}$ & ${\bf 63.7_{\pm 2.3}}$ \\ \hline
      DTD & $22.7_{\pm 1.3}$ & $27.8_{\pm 1.2}$ & ${\bf 35.6_{\pm 0.7}}$ \\ \hline
    \end{tabular}
    \vspace{2mm}
    \caption{Classification results for networks trained from scratch indicated as accuracy (\%).}
    \vspace{-1mm}
    \label{Table:accuracy_fromscratch}
\end{table}

\begin{figure}[t]
  \centering
  \includegraphics[width=\columnwidth]{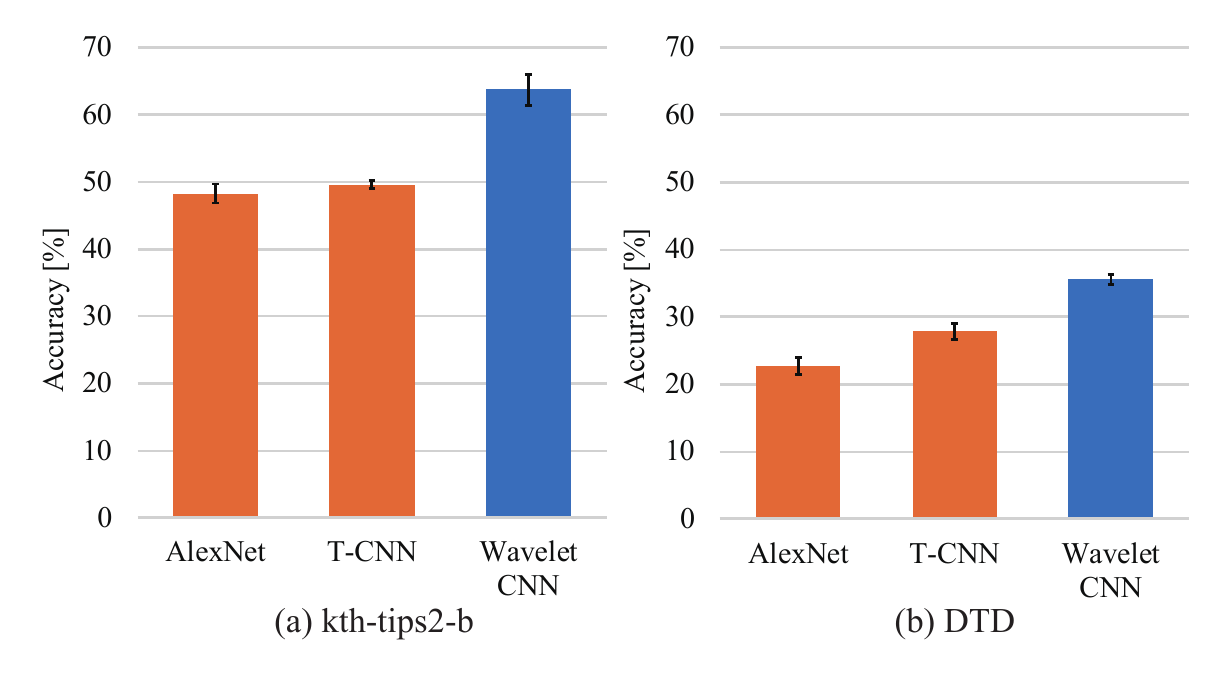}
  \vspace{-6mm}
  \caption{Classification results of (a) kth-tips2-b and (b) DTD for networks trained from scratch. We compared our models with AlexNet and T-CNN.}
  \vspace{-2mm}
  \label{Fig:accuracy_fromscratch}
\end{figure}

\paragraph{Training from scratch:}
Table~\ref{Table:comparison_between_levels} shows the results of our model with different levels of a multiresolution analysis.
For initialization of the parameters, we used a robust method for ReLU~\cite{He:2015}.
For both datasets, the network with 5-level decomposition performed the best, though the model with 4-level decomposition achieved almost the same accuracy as 5-level.
Figure~\ref{Fig:accuracy_fromscratch} and Table~\ref{Table:accuracy_fromscratch} compare our model with AlexNet~\cite{Krizhevsky:2012} and T-CNN~\cite{Andrearczyk:2016} using texture datasets to train each model from scratch.
Since the model with 5-level decomposition achieved the best accuracy in the previous experiment, we used this network in this and following experiments as well.
Since VGG networks tend to perform poorly due to over-fitting if trained from scratch, we used AlexNet as an example of conventional CNNs for this experiment.
For both datasets, our model performs better than AlexNet and T-CNN by a large margin.

\begin{figure}[t]
  \centering
  \includegraphics[width=\columnwidth]{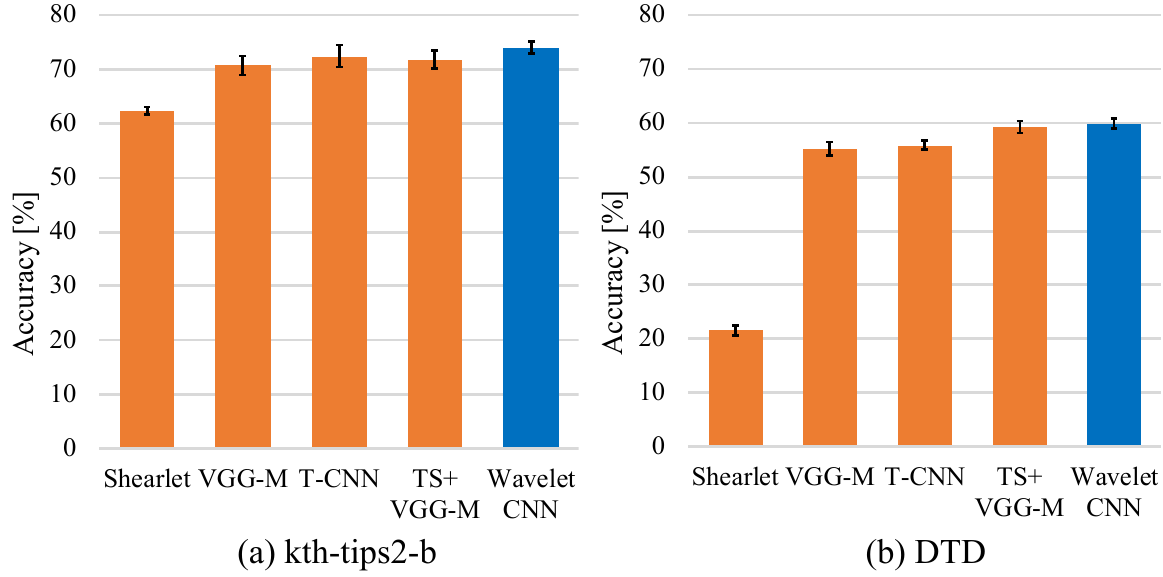}\vspace{-0.1cm}
  \vspace{-2mm}
  \caption{Classification results of (a) kth-tips2-b and (b) DTD for networks pre-trained with ImageNet 2012 dataset. We compared our model (wavelet CNN with 5-level decomposition) with shearlet transform, VGG-M, T-CNN, and TS+VGG-M.}
  \vspace{-2mm}
  \label{Fig:accuracy_finetuned}
\end{figure}

\paragraph{Training with fine-tuning:}
Figure~\ref{Fig:accuracy_finetuned} and Table~\ref{Table:accuracy_finetuned} show the classification rates using the networks pre-trained with the ImageNet 2012 dataset~\cite{Russakovsky:2015}.
%
% Since the model with 4-level decomposition achieved the best accuracy in the previous experiment, we used this network in this experiment as well.
%
We compared our model with a spectral approach using shearlet transform~\cite{Krishnan:2016}, VGG-M~\cite{Chatfield:2014}, T-CNN~\cite{Andrearczyk:2016}, and VGG-M using compact bilinear pooling~\cite{Gao:2016}.
For compact bilinear pooling, we compared our model only with Tensor Sketch (TS) since it worked the best in practice.
Our model again achieved the best performance for both datasets.
While the improvement for the \emph{DTD} dataset might be marginal (less than 1\%), as we show later, this performance is achieved with a significantly fewer parameters than other methods.

\begin{table}[t]
    \centering
    \scalebox{0.8}{
      \begin{tabular}{| l || l | l | l | l || l |} \hline
        & \raisebox{0.1cm}{Shearlet} & \raisebox{0.1cm}{VGG-M} & \raisebox{0.1cm}{T-CNN } & \raisebox{-0.085cm}{\shortstack{TS+ \\ VGG-M}} & \raisebox{-0.085cm}{\shortstack{Wavelet \\ CNN}} \rule[0mm]{0mm}{6mm}\\ \hline
        kth-tips2-b & $62.3_{\pm 0.8}$ & $70.7_{\pm 1.7}$ & $72.4_{\pm 2.1}$ & $71.8_{\pm 1.6}$ & ${\bf 74.0_{\pm 1.2}}$ \rule[0mm]{0mm}{3.5mm}\\ \hline
        DTD & $21.6_{\pm 0.9}$ & $55.2_{\pm 1.2}$ & $55.8_{\pm 0.8}$ & $59.3_{\pm 1.1}$ & ${\bf 59.8_{\pm 0.9}}$ \rule[0mm]{0mm}{3.5mm}\\ \hline
      \end{tabular}
    }
    \caption{Classification results for networks pre-trained with ImageNet indicated as accuracy (\%).}
    \label{Table:accuracy_finetuned}
\end{table}

\begin{figure*}[t]
  \vspace{-5mm}
  \centering
  \includegraphics[width=\textwidth]{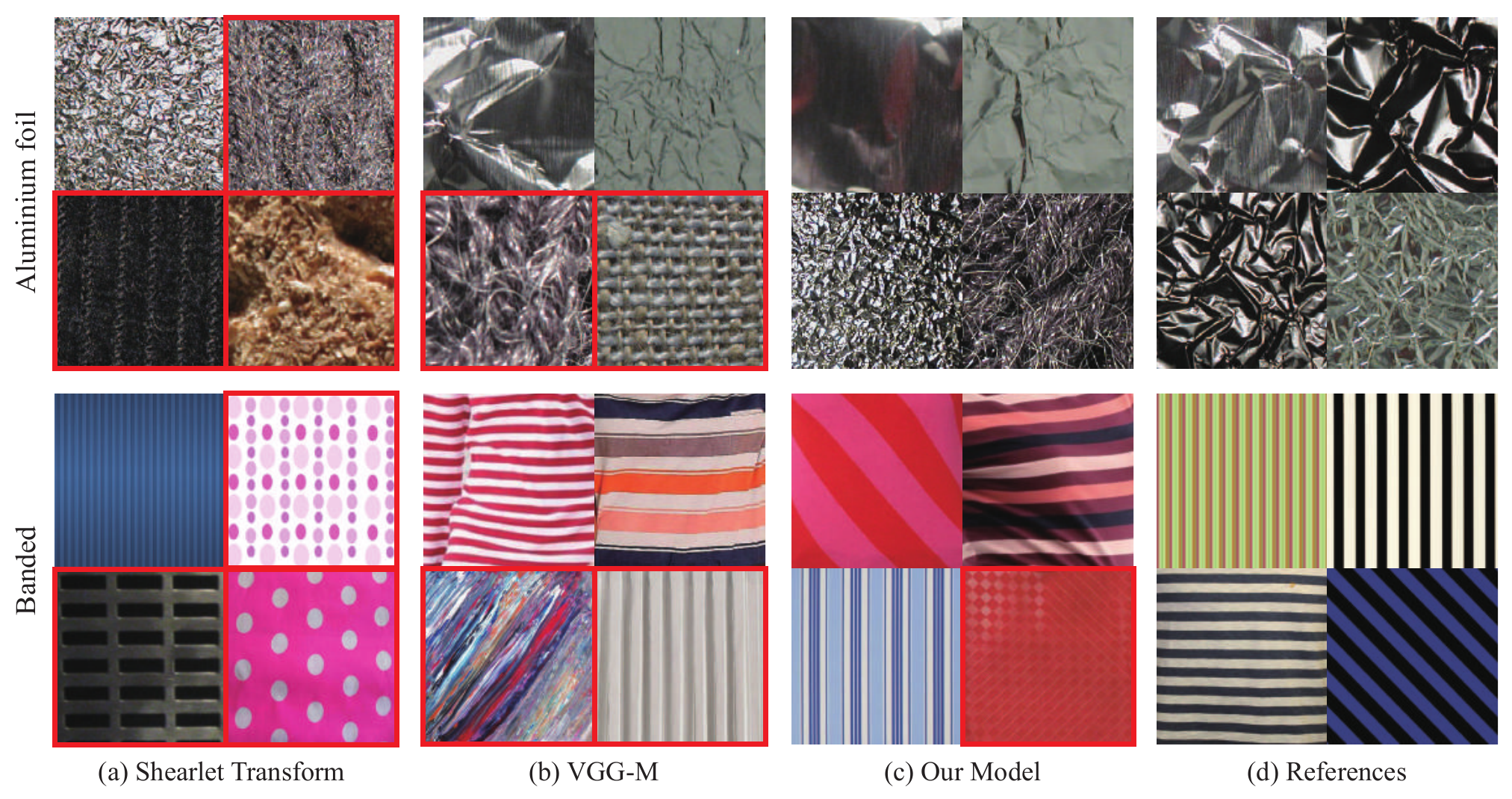}
  \vspace{-5mm}
  \caption{Some results classified by (a) shearlet transform, (b) VGG-M, (c) our model and (d) references. The images on the top row are extracted from \emph{kth-tips2-b} and the rests are extracted from \emph{DTD}. The images in red squares are wrongly classified images.}
  \label{Fig:results_TC}
\end{figure*}

\paragraph{Visual comparisons of classified images:}
Figure~\ref{Fig:results_TC} shows some extracted images for several classes in our experiments.
The images in the top row are from \emph{kth-tips2-b} dataset, while the images in the bottom row of Figure~\ref{Fig:results_TC} are from \emph{DTD} dataset.
A red square indicates a incorrectly classified texture.
We can visually confirm that a spectral approach (shearlet) is insensitive to the scale variation and extract detailed features, whereas a spatial approach (VGG-M) is insensitive to distortion.
For example, in \emph{Aluminium foil}, a shearlet transform can correctly ignore the scale of wrinkles, but VGG-M failed to classify such an image into the same class.
In \emph{Banded}, VGG-M classifies distorted lines into the correct class, but a shearlet transform could not recognize this line-like structure well.
Since our model is the combination of both approaches, it can assign texture images to the correct label in every variation above.

\subsection{Image Annotation}
The purpose of an image annotation task is to associate multiple labels with an image regarding to its content.
This task is more natural than single-label image classification because a natural image actually includes various objects.
The convolutional neural network - recurrent neural network (CNN-RNN) encoder-decoder model is a popular approach~\cite{Wang:2016,Jin:2016,Liu:2016} for this task.
In this model, a CNN encodes the image into a fixed length vector, and then it is fed into an RNN that decodes it into a list of tags.
The existing models share this concept and differ slightly in how the CNN and RNN relate to each other.

Recurrent image annotator (RIA)~\cite{Jin:2016} exploits image features output from the CNN as the RNN hidden states.
They focus on the order of a list of input tags and show that the rare-first order, which put rarer tags first based on their frequency, improves the performance.
Liu et al. proposed semantically regularized CNN-RNN (S-CNN-RNN)~\cite{Liu:2016} where the CNN model is regularized by semantic concepts which serve as strong deep supervision to guide the learning of the CNN layers.
The prediction layer of the CNN in this model is also used as the RNN initial states.
Both models use VGG-16~\cite{Simonyan:2014} as the CNN and the long short-term memory (LSTM)~\cite{Hochreiter:1997} as RNN.
In our experiment, we compared our model with RIA.

\paragraph{Datasets:}
We used two benchmark image annotation datasets: \emph{IAPR-TC12}~\cite{Michael:2007} and \emph{Microsoft COCO}~\cite{Lin:2014}.
The \emph{IAPR-TC12} dataset contains 20,000 images of natural scenes with text captions in several languages.
To use this dataset for image annotation, it can be arranged by extracting common nouns in accord with the previous work~\cite{Makadia:2008}.
This process results in a vocabulary size of 291.
Training used 17,665 images while the remaining are used for testing.
The \emph{Microsoft COCO} (MS-COCO) dataset contains 82,783 training images and 40,504 testing images.
Following the previous works~\cite{Wang:2016,Liu:2016}, we employed 80 object annotations as labels.

\paragraph{Training Details:}
We replaced VGG-16 in RIA by a wavelet CNN with 5-level decomposition.
For LSTM, the dimension of both hidden states and the input is set to 1024 and the number of hidden layer is 1.
When training, the hidden state and the cell state are initialized by image features from CNN and zero respectively.
Additionally, since original RIA uses 4096 dimensional output from the last fully-connected layer of VGG-16 as image features, we add a 2048 dimensional fully connected layer to our model just after an average pooling layer and exploit the output from this layer as image features.
Even though this additional layer increases the number of trainable parameters in our model, our model still has only 18.3 millions parameters while VGG-16 has 138.4 millions parameters.
For the order of input tags, we use the rare-first order following the original paper~\cite{Jin:2016}.

\begin{table}[t]
    \centering
    \scalebox{0.85}{
      \begin{tabular}{| l || l | l | l || l | l | l |} \hline
        & {C-P} & {C-R} & {C-F1} & {O-P} & {O-R} & {O-F1} \\ \hline
        VGG-16~\cite{Simonyan:2014} & 22.97 & 27.39 & 24.99 & 33.87 & 34.93 & 34.40 \\ \hline
        Wavelet CNN & \bf 29.01 & \bf 30.62 & \bf 29.79 & \bf 37.43 & \bf 37.66 & \bf 37.54 \\ \hline
      \end{tabular}
    }
    \caption{Annotation results of RIAs for IAPR-TC12.}
    \vspace{-2mm}
    \label{Table:annotaions_iapr}
\end{table}

\begin{table}[t]
    \centering
    \scalebox{0.85}{
      \begin{tabular}{| l || l | l | l || l | l | l |} \hline
        & {C-P} & {C-R} & {C-F1} & {O-P} & {O-R} & {O-F1} \\ \hline
        VGG-16~\cite{Simonyan:2014} & 51.55 & 45.60 & 48.49 & 57.94 & 51.92 & 54.77 \\ \hline
        Wavelet CNN & \bf 53.17 & \bf 46.69 & \bf 49.72 & \bf 58.68 & \bf 52.04 & \bf 55.16 \\ \hline
      \end{tabular}
    }
    \caption{Annotation results of RIAs for Microsoft COCO.}
    \vspace{-2mm}
    \label{Table:annotaions_MS}
\end{table}

\paragraph{Results:}
We used per-class and overall metrics including precision (C-P and O-P), recall (C-R and O-R) and F1 score (C-F1 and O-F1) as evaluation metrics.
Per-class metrics take the average over all classes while overall metrics take the average over all test images.
As RIA can produce annotations in arbitrary length, we used the arbitrary-length results to compare.
Table~\ref{Table:annotaions_iapr} and Table~\ref{Table:annotaions_MS} show the results using RIA models with VGG-16 and the wavelet CNN for IAPR-TC12 and MS-COCO.
For IAPR-TC12, RIA with our model obtained much better results than original RIA.
The improvement for MS-COCO was marginal in comparison.
However, all these results are obtained with a significantly fewer number of parameters; the number of parameters of wavelet CNN is more than seven times smaller than that of VGG-16.

\begin{figure*}[t]
  \vspace{-5mm}
  \centering
  \includegraphics[width=\textwidth]{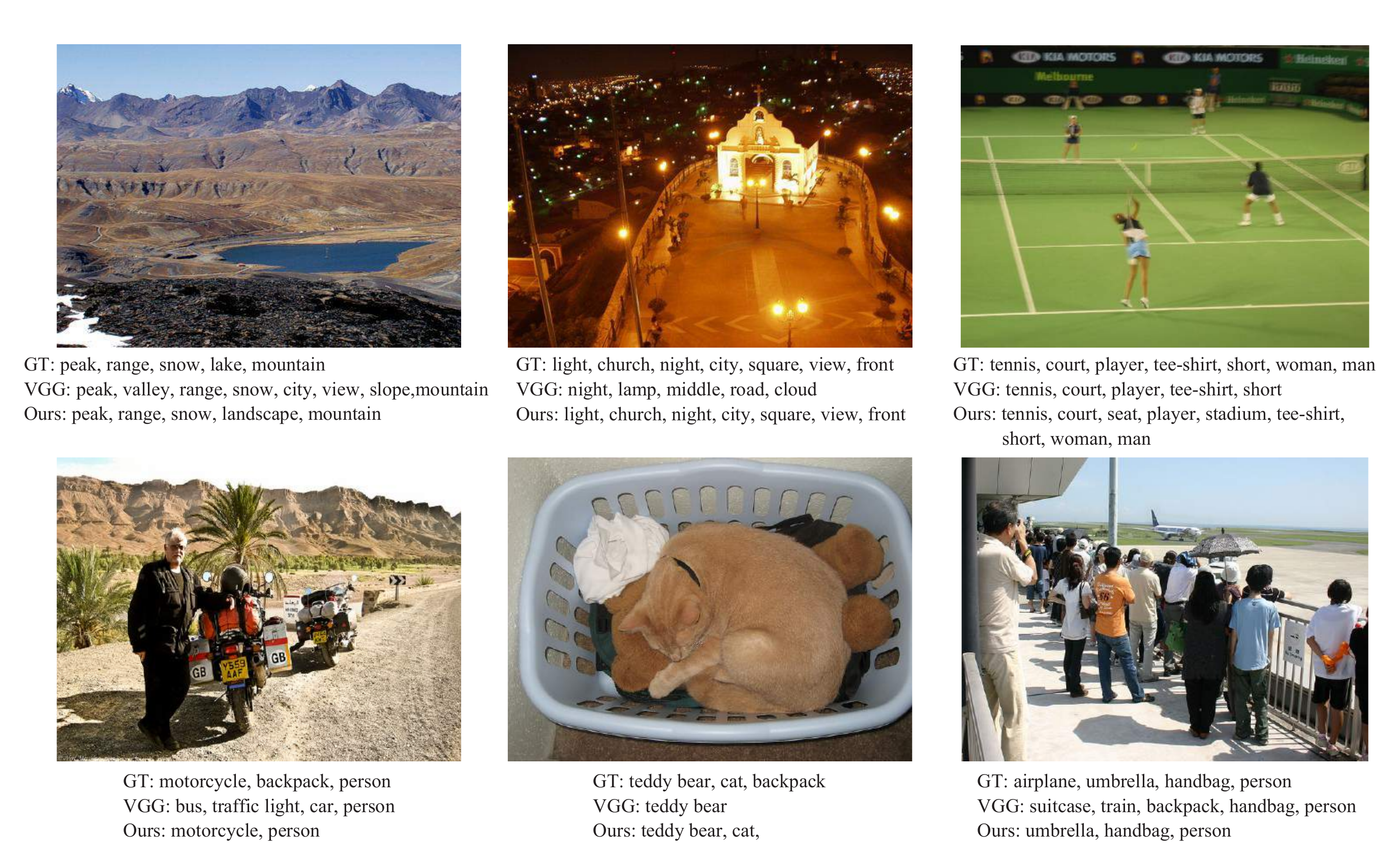}
  \vspace{-5mm}
  \caption{Some results of image annotaion. The images on the top row are extracted from \emph{IAPR-TC12} and the rests are extracted from \emph{Microsoft COCO}. GT, VGG and ours show the ground-truth annotaions, the predictions of RIA with VGG-16 and the predictions of RIA with wavelet CNN, respectively.}
  \label{Fig:results_tags}
\end{figure*}

Figure~\ref{Fig:results_tags} shows some results in image annotation. The images in the top row are from IAPR-TC12 while the images in the bottom row are from MS-COCO.
GT indicates the ground-truth annotations, and they are organized in the rare-first order.
VGG and ours show the results of RIA with VGG-16 and our model, where the order of the predictions is preserved as RNN output.

\subsection{Number of parameters}
\begin{figure}[t]
  \vspace{-3mm}
  \centering
  \includegraphics[width=\columnwidth]{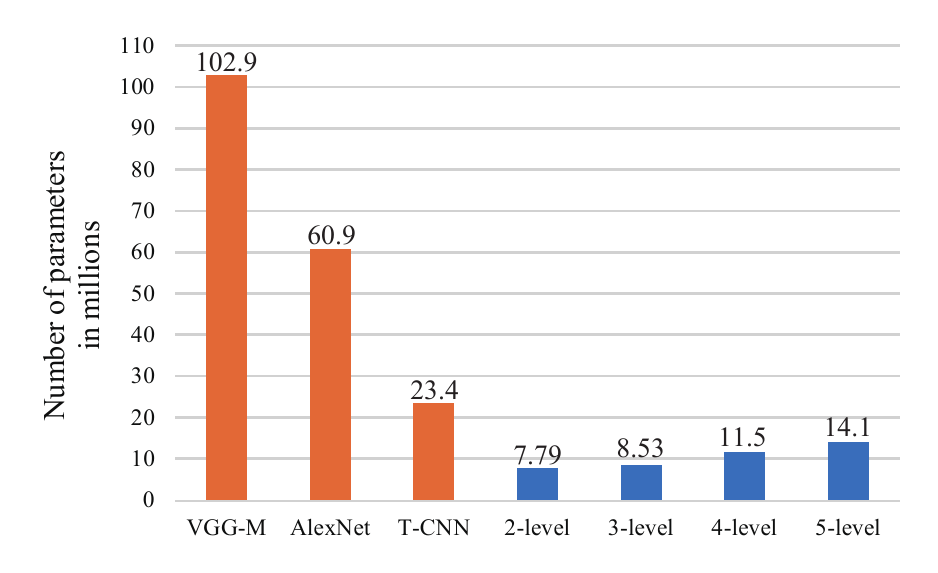}\vspace{-0.4cm}
  \vspace{-3mm}
  \caption{The number of trainable parameters in millions. Our model, even with 5-level of multiresolution analysis, has a fewer parameters than any other competing models we tested.}
  \vspace{-2mm}
  \label{Fig:parameters}
\end{figure}

To assess the complexity of each model, we compared the number of trainable parameters such as weights and biases for classification to 1000 classes (Figure~\ref{Fig:parameters}).
Conventional CNNs such as VGG-M and AlexNet have a large number of parameters while their depth is a little shallower than our proposed model.
Even compared to T-CNN, which aims at reducing the model complexity, the number of parameters in our model with 5-level decomposition is about the half.
We also remind that our model achieved higher accuracy than T-CNN does in texture classification.

This result confirms that our model achieves better results with a significantly reduced number of parameters than existing models.
The memory consumption of each Caffe model is: 392~MB (VGG-M), 232~MB (AlexNet), 89.1~MB (T-CNN), and 53.9~MB (Ours).
The small number of parameters generally suppresses over-fitting of the model for small datasets.

%% file: discussion.tex
%!TeX root=paper.tex

%%%%%%%%%%%%%%%%%%%%%%%%%%%%%%%%%%%%%%%%%%%%%%%%%%%%%%%%%%%%%%%
\section{Discussion}\label{Sec:Discussion}
%%%%%%%%%%%%%%%%%%%%%%%%%%%%%%%%%%%%%%%%%%%%%%%%%%%%%%%%%%%%%%%
\paragraph{Application to more general tasks:}
We applied our model to two challenging tasks; texture classification and image annotation.
However, since we do not assume anything regarding the input, our model is not necessarily restricted to these tasks.
For example, we experimented training a wavelet CNN with 5-level decomposition and AlexNet with the ImageNet 2012 dataset from scratch to perform image classification.
Our model obtained the accuracy of 59.4\% whereas AlexNet resulted in 57.1\%.
We should remind that the number of parameters of our model is about four times smaller than that of AlexNet (Figure~\ref{Fig:parameters}).
Our model is thus suitable also for image classification with smaller memory footprint.
Other applications such as image recognition and object detection with our model should be similarly possible.

\paragraph{$\mbox{\boldmath $L_p$}$ pooling:}
An interesting generalization of max and average pooling is $L_p$ pooling~\cite{Boureau:2010,Sermanet:2012}.
The idea of $L_p$ pooling is that max pooling can be thought as computing $L_{\infty}$ norm, while average pooling can be considered as computing $L_{1}$ norm.
In this case, Equation~\ref{Eqn:Pooling2} cannot be written as linear convolution anymore due to non-linear transformation in norm calculation.
Our overall formulation, however, is not necessarily limited to a multiresolution analysis either; we can just replace downsampling part by corresponding norm computation to support $L_p$ pooling.
This modification however will not retain all the frequency information of the input as it is no longer a multiresolution analysis.
We focused on average pooling as it has a clear connection to a multiresolution analysis.

\paragraph{Limitations:}
We designed wavelet CNNs to put each high frequency part between layers of the CNN.
Since our network has four layers to reduce the size of feature maps, the maximum decomposition level is restricted to five.
This design is likely to be less ideal since we cannot tweak the decomposition level independently from the depth (thereby the number of trainable parameters) of the network.
A different network design might make this separation of hyper-parameters possible.

Wavelet CNNs achieved the best accuracy for both training from scratch and with fine-tuning for texture classification.
For the performance with fine-tuning, however, our model outperforms other methods by a slight margin especially for \emph{DTD}, albeit with a significantly smaller number of parameters.
We speculated that it is partially because pre-training with the ImageNet 2012 dataset is simply not appropriate for texture classification.
An exact reasoning of failure cases for texture classification, however, is generally difficult for any neural network models, and our model is not an exception.
%

%% file: conclusion.tex
%!TeX root=paper.tex

%%%%%%%%%%%%%%%%%%%%%%%%%%%%%%%%%%%%%%%%%%%%%%%%%%%%%%%%%%%%%%%
\section{Conclusion}\label{Sec:Conclusion}
%%%%%%%%%%%%%%%%%%%%%%%%%%%%%%%%%%%%%%%%%%%%%%%%%%%%%%%%%%%%%%%
%
We presented a novel CNN architecture which incorporates a spectral analysis into CNNs.
We showed how to reformulate convolution and pooling layers in CNNs into a generalized form of filtering and downsampling.
This reformulation shows how conventional CNNs perform a limited version of multiresolution analysis, which then allows us to integrate multiresolution analysis into CNNs as a single model called wavelet CNNs.
We demonstrated that our model achieves better accuracy for texture classification and image annotation with smaller number of trainable parameters than existing models.
In particular, our model outperformed all the existing models with significantly more trainable parameters by a large margin when we trained each model from scratch.
A wavelet CNN is a general learning model and applications to other problems are interesting future works.
Finally, for wavelet transform in our model, we use fixed-weight kernels as Haar wavelet.
It is interesting to explore how wavelet kernels themselves can be trained in the end-to-end learning framework.